\documentclass[11pt,a4paper]{article}
\usepackage[hyperref]{naaclhlt2019}
\usepackage{times}
\usepackage{latexsym}
\usepackage{graphicx}
\usepackage{subcaption}
\usepackage{wrapfig}
\usepackage{overpic}
\usepackage{hyperref}
\usepackage{url}
\usepackage{mathtools}
\usepackage{amssymb}
\usepackage{bm}
\usepackage[warn]{textcomp}
\usepackage[inline]{enumitem}
\usepackage{comment}
\usepackage{booktabs}
\usepackage{multirow}
\usepackage{tikz}
\usepackage{stfloats}

\aclfinalcopy 
\def\aclpaperid{2052} 

\newcommand*{\Resize}[2]{\resizebox{#1}{!}{$#2$}}

\newcommand\blfootnote[1]{%
  \begingroup
  \renewcommand\thefootnote{}\footnote{#1}%
  \addtocounter{footnote}{-1}%
  \endgroup
}

\DeclareMathOperator{\mean}{mean}
\usetikzlibrary{math}

\title{Adversarial Decomposition of Text Representation}

\author{
    Alexey Romanov, Anna Rumshisky, Anna Rogers and David Donahue \\
    Department of Computer Science \\
    University of Massachusetts Lowell  \\
    Lowell, MA 01854\\ 
    {\tt \{aromanov,arum,arogers\}@cs.uml.edu} \\
    {\tt david\_donahue@student.uml.edu}
}

\date{}

\begin{document}
\maketitle
\begin{abstract}

In this paper, we present a method for adversarial decomposition of text representation. This method can be used to decompose a representation of an input sentence into several independent vectors, each of them responsible for a specific aspect of the input sentence. We evaluate the proposed method on two case studies: the conversion between different social registers and diachronic language change. We show that the proposed method is capable of fine-grained controlled change of these aspects of the input sentence. It is also learning a continuous (rather than categorical) representation of the style of the sentence, which is more linguistically realistic. The model uses adversarial-motivational training and includes a special motivational loss, which acts opposite to the discriminator and encourages a better decomposition. Furthermore, we evaluate the obtained meaning embeddings on a downstream task of paraphrase detection and show that they significantly outperform the embeddings of a regular autoencoder.

\end{abstract}

\blfootnote{Accepted at NAACL 2019}

\section{Introduction}

Despite the recent successes in using neural models for representation learning for natural language text, learning a meaningful representation of input sentences remains an open research problem. 
A variety of approaches, from sequence-to-sequence models that followed the work of~\citet{sutskever2014sequence} to the more recent proposals~\citep{arora2016simple,nangia2017repeval,conneau2017supervised,logeswaran2018efficient,subramanian2018learning,cer2018universal} share one common drawback. Namely, all of them encode the input sentence into just \textit{one} single vector of a fixed size. 
One way to bypass the limitations of a single vector representation is to use an attention mechanism~\citep{bahdanau2014neural,vaswani2017attention}. 
We propose to approach this problem differently and design a method for adversarial decomposition of the learned input representation into multiple components. Our method encodes the input sentence into \textit{several} vectors, where each vector is responsible for a specific aspect of the sentence. 

In terms of learning different separable components of input representation,
our work most closely relates to the style transfer work, which has been
applied to a variety of different aspects of language, from diachronic language differences~\citep{xu2012paraphrasing} to authors' personalities~\citep{lipton2015generative} and even sentiment~\citep{hu2017toward,fu2018style}.
The style transfer work effectively relies on the more classical distinction between \textbf{meaning} and \textbf{form} \citep{Saussure_1959_Course_in_general_linguistics}, which accounts for the fact that multiple surface realizations are possible for the same meaning.  
For simplicity, we will use this terminology throughout the rest of the paper.

Consider encoding an input sentence into a meaning vector and a form vector. This enables a controllable change of meaning or form by a simple change applied to these vectors. For example, we can encode two sentences written in two different styles, then swap the form vectors while leaving the meaning vectors intact. We can then generate new unique sentences with the original meaning, but written in a different style. 

We propose a novel model for this type of decomposition based on adversarial-motivational training, GAN architecture~\citep{goodfellow2014generative} and adversarial autoencoders~\citep{makhzani2015adversarial}. In addition to the adversarial loss, we use a special motivator~\citep{albanie2017stopping}, which, in contrast to the discriminator, is used to provide a motivational loss to encourage better decomposition of the meaning and the form.
All the code is available on  GitHub~\footnote{\url{https://github.com/text-machine-lab/adversarial_decomposition}}.

We evaluate the proposed methods for learning separate aspects of input representation in the following case studies: 
\begin{enumerate}[noitemsep]
\item Diachronic language change. Specifically, we consider the Early Modern English (e.g. \textit{What would she have?}) and the contemporary English (\textit{ What does she want?}).
\item Social register~\citep{HallidayMcIntoshEtAl_1968}, i.e. subsets of language appropriate in a given context or characteristic of a certain group of speakers.  Social registers include formal vs informal language, the language used in different genres (e.g., fiction vs. newspapers vs. academic texts), different dialects, and literary idiostyles. We experiment with the titles of scientific papers vs. newspaper articles.
\end{enumerate}

\section{Related work}
\label{sec:related_work}


As mentioned above, the most relevant previous work comes from research on style transfer\footnote{The term ``style'' is not entirely appropriate here, but in NLP it is often used in work on any kind of form change while preserving meaning, from translation to changing sentiment polarity.}. It can be divided into two groups:
\begin{enumerate}[noitemsep]
    \item Approaches that aim to generate text in a given form. For example, the task may be to produce just any verse as long as it is in the ``style'' of the target poet.
    \item Approaches that aim to induce a change in either the ``form'' or the ``meaning'' of an utterance. For example, ``Good bye, Mr. Anderson.'' can be transformed to ``Fare you well, good Master Anderson''~\citep{xu2012paraphrasing}).    
\end{enumerate}

An example of the first group is the work of ~\citet{potash2015ghostwriter}, who trained several separate networks on verses by different hip-hop artists. An LSTM network successfully generated verses that were stylistically similar to the verses of the target artist (as measured by cosine distance on tf-idf vectors). More complicated approaches use language models that are conditioned in some way. For example, \citet{lipton2015generative} produced product reviews with a target rating by passing the rating as an additional input at each timestep of an LSTM model. \citet{tang2016context} generated reviews not only with a given rating but also for a specific product. At each timestep a special context vector was provided as input, gated so as to enable the model to decide how much attention to pay to that vector and the current hidden state. \citet{li2016persona} used ``speaker'' vectors as an additional input to a conversational model, improving consistency of dialog responses. 
Finally, \citet{ficler2017controlling} performed an extensive evaluation of conditioned language models based on ``content'' (theme and sentiment) and ``style'' (professional, personal, length, descriptiveness). Importantly, they showed that it is possible to control both ``content'' and ``style'' simultaneously.

Work from the second group can further be divided into two clusters by the nature of the training data: parallel aligned corpora, or non-aligned datasets. The aligned corpora enable approaching the problem of form shift  as a paraphrasing or machine translation problem. \citet{xu2012paraphrasing} used statistical and dictionary-based systems on a dataset of original plays by Shakespeare and their contemporary translations. \citet{carlson2017zero} trained an LSTM network on 33 versions of the Bible. \citet{jhamtani2017shakespearizing} used a Pointer Network~\citep{vinyals2015pointer}, an architecture that was successfully applied to a wide variety of tasks~\citep{merity2016pointer,gulcehre2016pointing,potash2017here}, 
to enable direct copying of the input tokens to the output. All these works use BLEU~\citep{papineni2002bleu} as the main, or even the only evaluation measure. This is only possible in cases where a parallel corpus is available. 

Recently, new approaches that do not require a parallel corpora were developed in both computer vision (CV)~\citep{zhu2017unpaired} and NLP. \citet{hu2017toward} succeeded in changing tense and sentiment of sentences with a two steps procedure based on a variational auto-encoder (VAE)~\citep{kingma2013auto}. After training a VAE, a discriminator and a generator are trained in an alternate manner, where the discriminator tries to correctly classify the target sentence attributes. 
A special loss component forces the hidden representation of the encoded sentence to not have any information about the target sentence attributes. 
\citet{mueller2017sequence} used a VAE to produce a hidden representation of a sentence, and then modify it to match the desired form. Unlike ~\citet{hu2017toward}, they do not separate the form and meaning embeddings. \citet{shen2017style} applied a GAN to align the hidden representation of sentences from two corpora and forced them not to have any information about the form an via adversarial loss. During the decoding, similarly to \citet{lipton2015generative}, special ``style'' vectors are passed to the decoder at every timestep to produce a sentence with the desired properties. The model is trained using the Professor-Forcing algorithm~\citep{lamb2016professor}. \citet{kim2017adversarially} worked directly on hidden space vectors that are constrained with the same adversarial loss instead of outputs of the generator, and use two different generators for different ``styles''. Finally, ~\citet{fu2018style} generate sentences with the target properties using an adversarial loss, similarly to~\citet{shen2017style} and~\citet{kim2017adversarially}. 


\paragraph*{Comparison with previous work}

In contrast to the proposals of~\citet{xu2012paraphrasing},~\citet{carlson2017zero},~\citet{jhamtani2017shakespearizing}, our solution does not require a parallel corpus. Unlike the model by~\citet{shen2017style}, our model works directly on representations of sentences in the hidden space.

Most importantly, in contrast to the proposals by~\citet{mueller2017sequence},~\citet{hu2017toward},~\citet{kim2017adversarially},~\citet{fu2018style}, our model produces a representation for both meaning and form and does not treat the form as a categorical (in the vast majority of works, binary) variable\footnote{Although the form was represented as dense vectors in previous work, it is still just a binary feature, as they use a single pre-defined vector for each form, with all sentences of the same form assigned the same form vector.}.

Treating meaning and form not as binary/categorical, but continuous variables
is more consistent with the reality of language use, since there are different degrees of overlap between the language used by different registers or in different diachronic slices. Indeed, language change is gradual, and the acceptability of expressions in a given register also forms a continuum, so one expects a substantial overlap between the grammar and vocabulary used, for example, on Twitter and by New York Times.
To the best of our knowledge, this is the first model that considers linguistic form in the task of text generation as a continuous variable.

A significant consequence of learning a continuous representation for form is that it allows the model to work with a large, and potentially infinite, number of forms. Note that in this case the locations of areas of specific forms in the vector form space would reflect the similarity between these forms. For example, the proposed model could be directly applied to the authorship attribution problem: each author would have their own area in the form space, their proximity should mirror the similarity in writing style. Preliminary experiments on this are reported in~\autoref{sec:multiple_forms}.

\section{Formulation}

Let us formulate the problem of decomposition of text representation on an example of controlled change of linguistic form and conversion of Shakespeare plays in the original Early Modern to contemporary English. 
Let $\bm{X}^a$ be a corpus of texts $\bm{x}^a_i \in \mathcal{X}^a$ in Early Modern English $\bm{\mathrm{f}^a} \in \mathcal{F}$, and $\bm{X}^b$ be a corpus of texts $\bm{x}^b_i \in \mathcal{X}^b$ in modern English $\bm{\mathrm{f}^b} \in \mathcal{F}$. We assume that the texts in both $\bm{X}^a$ and $\bm{X}^b$ have the same distribution of meaning $\bm{\mathrm{m}} \in \mathcal{M}$. The form $\bm{\mathrm{f}}$, however, is different and generated from a mixture of two distributions: 
$$
 \bm{\mathrm{f}}_i = \alpha^a_i p(\bm{\mathrm{f}}^a) + \alpha^b_i p(\bm{\mathrm{f}}^b) 
$$
 
where $\bm{\mathrm{f}}^a$ and $\bm{\mathrm{f}}^b$ are two different languages (Early Modern and contemporary English). Intuitively, we say that a sample $\bm{x}_i$ has the form $\bm{\mathrm{f}^a}$ if $\alpha^a_i > \alpha^b_i$, and it has the form $\bm{\mathrm{f}^b}$ if $\alpha^b_i > \alpha^a_i$. 

The goal of dissociation meaning and form is to learn two encoders $E_{\bm{\mathrm{m}}}:\mathcal{X} \rightarrow \mathcal{M} $ and
$E_{\bm{\mathrm{f}}}:\mathcal{X} \rightarrow \mathcal{F} $ for the meaning and form correspondingly,
and the generator $G:\mathcal{M}, \mathcal{F} \rightarrow  \mathcal{X}$ such that 
$$
    \small
   \forall j \in \{a,b\}, \forall k \in \{a,b\} : G(E_{\bm{\mathrm{m}}}(\bm{x}^k),E_{\bm{\mathrm{f}}}(\bm{x}^j)) \rightarrow \mathcal{X}^j 
$$
The form of a generated sample depends exclusively on the provided $\bm{\mathrm{f}}_j$ and can be in the same domain for two different $\bm{\mathrm{m}}_u$ and $\bm{\mathrm{m}}_v$ from two samples from different domains $\mathcal{X}^a$ and $\mathcal{X}^b$.

Note that, in contrast to the previous proposals, the form $\bm{\mathrm{f}}$ is not a categorical variable but a continuous vector. This enables fine-grained controllable change of form: the original form $\bm{\mathrm{f}}_i$ is changed to reflect the form of the specific target sentence $\bm{\mathrm{f}}_j$ with its own unique $\alpha^a$ and $\alpha^b$ while preserving the original meaning $\bm{\mathrm{m}}_i$. 

An important caveat concerns the core assumption of the similar meaning distribution in the two corpora, which is also made in all other works reviewed in Section~\ref{sec:related_work}. It limits the possible use of this approach to cases where the distributions are in fact similar (i.e. comparable corpora are available; note that they do not have to be parallel). It does not apply to many cases that could be analyzed in terms of meaning and form. For example, books for children and scholarly papers are both registers, they have their own form (i.e. specific subsets of linguistic means and structure conventions) -- but there is little overlap in the content. This would make it hard even for a professional writer to turn a research paper into a fairy tale.

\section{Method description}

\label{sec:method_description}
\begin{figure}
    \begin{overpic}[width=0.8\linewidth, trim=0.2cm 0.2cm 0.2cm 0.2cm, clip]{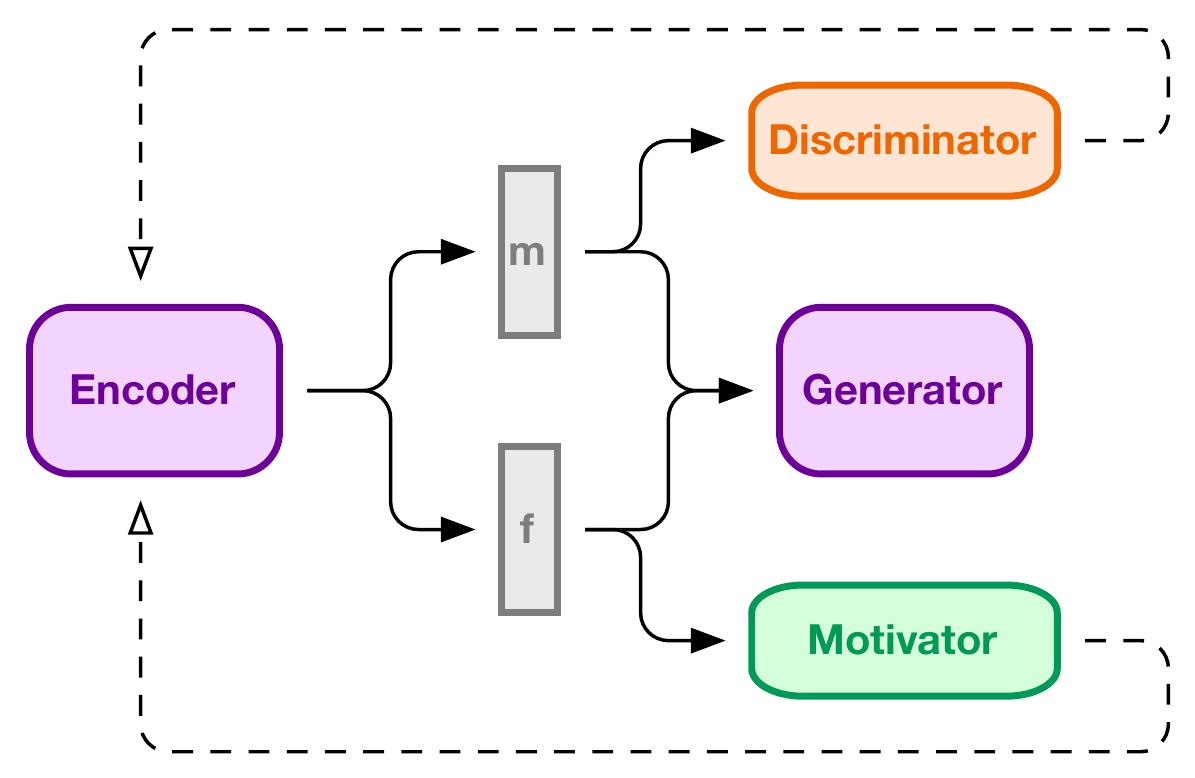}
        \put(102,57){\mbox{$L_\text{adv}$}}
        \put(102,3){\mbox{$L_\text{motiv}$}}
    \end{overpic}
    \caption{Overview of ADNet. $Encoder$ encodes the inputs sentences into two latent vectors $\bm{\mathrm{m}}$ and $\bm{\mathrm{f}}$. The $Generator$ takes them as the input and produces the output sentence. During the training, the $Discriminator$ is used for an adversarial loss that forces $\bm{\mathrm{m}}$ to not carry any information about the form, and the $Motivator$ is used for a motivational loss that encourages $\bm{\mathrm{f}}$ to carry the information about the form.
    } 
    \label{fig:model_zzz}
\end{figure}

Inspired by~\citet{makhzani2015adversarial}, \citet{kim2017adversarially}, and~\citet{albanie2017stopping}, we propose \mbox{ADNet}, a new model for adversarial decomposition of text representation (\autoref{fig:model_zzz}).

Our solution is based on a widely used sequence-to-sequence framework~\citep{sutskever2014sequence} and consists of four main parts. The encoder $E$ encodes the input sequence $\bm{x}$ into two latent vectors $\bm{\mathrm{m}}$ and $\bm{\mathrm{f}}$ which capture the meaning and the form of the sentence correspondingly. The generator $G$ then takes these two vectors as the input and produces a reconstruction of the original input sequence $\hat{\bm{x}}$.

The encoder and generator by themselves will likely not achieve the dissociation of the meaning and form. We encourage this behavior in a way similar to Generative Adversarial Networks (GANs)~\citep{goodfellow2014generative}, which had an overwhelming success the past few years as a way to enforce a specific distribution and characteristics on the output of a model. 

Inspired by the work of~\citet{albanie2017stopping} and the principle of ``carrot and stick"~\citep{safire1995on}, in contrast to the majority of work that promotes purely adversarial approach~\citep{goodfellow2014generative,shen2017style,fu2018style,zhu2017unpaired}, we propose two additional components, the discriminator $D$ and the motivator $M$ to force the model to learn the dissociation of the meaning and the form. Similarly to a regular GAN model, the adversarial discriminator $D$ tries to classify the form $\bm{\mathrm{f}}$ based on the latent meaning vector $\bm{\mathrm{m}}$, and the encoder $E$ is penalized to make this task as hard as possible. 

Opposed to such vicious behaviour,
the motivator $M$ tries to classify the form based on the latent form vector $\bm{\mathrm{f}}$, as it should be done, and encourages the encoder $E$ to make this task as simple as possible. 
We could apply the adversarial approach here as well and force the distribution of the form vectors to fit a mixture of Gaussians (in this particular case, a mixture of two Guassians) with another discriminator, as it is done by~\citet{makhzani2015adversarial}, but we opted for the ``dualistic'' path of two complimentary forces. 


\subsection{Encoder-Decoder}
Both the encoder $E$ and the generator $G$ are neural networks. Gated Recurrent Unit (GRU)~\citep{chung2014empirical} is used for $E$ to encode the input sentence $\bm{x}$ into a hidden vector 
$$\bm{h} = \text{GRU}(\bm{x})$$

The vector $\bm{h}$ then passes through two different fully connected layers to produce the latent vectors of the form and the meaning of the input sentence:
\begin{align*}
    \bm{\mathrm{m}} &= \tanh(\bm{W_m} \bm{h} + \bm{b}_m) \\ 
    \bm{\mathrm{f}} &= \tanh(\bm{W_f} \bm{h} + \bm{b}_f) 
\end{align*}

We use $\bm{\theta}_E$ to denote the parameters of the encoder $E$: $\bm{W_m}$, $\bm{b}_m$, $\bm{W_f}$, $\bm{b}_f$, and the parameters of the GRU unit.

The generator $G$ is also modelled with a GRU unit. The generator takes as input the meaning vector $\bm{\mathrm{m}}$ and the form vector $\bm{\mathrm{f}}$, concatenates them, and passes trough a fully-connected layer to obtain a hidden vector $\bm{z}$ that represents both meaning and form of the original input sentence:
$$
\bm{z} = \tanh(\bm{W}_z [\bm{\mathrm{m}};\bm{\mathrm{f}}] + \bm{b}_m) 
$$
After that, we use a GRU unit to generate the output sentence as a probability distribution over the vocabulary tokens:
$$ p(\hat{\bm{x}}) = \prod_{t=1}^T p(\hat{\bm{x}}_t|\bm{z},\hat{\bm{x}}_1, \dotsc , \hat{\bm{x}}_{t-1}) $$

We use $\bm{\theta}_G$ to denote the parameters of the generator $G$: $\bm{W}_z$, $\bm{b}_m$, and the parameters of the used GRU. The encoder and generator are trained using the standard reconstruction loss:
$$
\Resize{0.98\linewidth}{
    \mathcal{L}_{\text{rec}}(\bm{\theta}_E,\bm{\theta}_G) =\mathbb{E}_{\bm{x} \sim \bm{X}^a} [-\log p(\hat{\bm{x}}|\bm{x})] + \mathbb{E}_{\bm{x} \sim \bm{X}^b} [-\log p(\hat{\bm{x}}|\bm{x})] 
}
$$

\subsection{Discriminator}
The representation of the meaning $\bm{\mathrm{m}}$ produced by the encoder $E$ should not contain any information about the form $\bm{\mathrm{f}}$. We achieve this by using an adversarial approach. First, we train a discriminator $D$, consisting of several fully connected layers with \verb|ELU| activation function~\citep{clevert2015fast} between them, to predict the form $\bm{\mathrm{f}}$ of a sentence by its meaning vector:
$$ \hat{\bm{f}}_D = D(\bm{\mathrm{m}}) $$ 
where $\hat{\bm{f}}$ is the score (logit) reflecting the probability of the sentence $\bm{x}$ to belong to one of the  form domains.

Motivated by the Wasserstein GAN~\citep{arjovsky2017wasserstein}, we use the following loss function instead of the standard cross-entropy:
$$
\Resize{0.98\linewidth}{
   \mathcal{L}_D(\bm{\theta}_D) = \mathbb{E}_{\bm{x} \sim \bm{X}^a} [D(E_{\bm{\mathrm{m}}}(\bm{x}))] - \mathbb{E}_{\bm{x} \sim \bm{X}^b} [D(E_{\bm{\mathrm{m}}}(\bm{x}))] 
   }
$$

Thus, a successful discriminator will produce negative scores $\hat{\bm{f}}$ for sentences from $\bm{X}^a$ and positive scores for sentences from $\bm{X}^b$. This discriminator is then used in an adversarial manner to provide a learning signal for the encoder and force dissociation of the meaning and form by maximizing $\mathcal{L}_D$:
$$ \mathcal{L}_\text{adv}(\bm{\theta}_E) = -\lambda_\text{adv} \mathcal{L}_D $$ 
where $\lambda_\text{adv}$ is a hyperparameter reflecting the strength of the adversarial loss. Note that this loss applies to the parameters of the encoder.

\subsection{Motivator}

Our experiments showed that the discriminator $D$ and the adversarial loss $\mathcal{L}_\text{adv}$ by themselves are sufficient to force the model to dissociate the form and the meaning. However, in order to achieve a better dissociation, we propose to use a motivator $M$~\citep{albanie2017stopping} and the corresponding motivational loss. Conceptually, this is the opposite of the adversarial loss, hence the name. As the discriminator $D$, the motivator $M$ learns to classify the form $\bm{\mathrm{f}}$ of the input sentence. However, its input is not the meaning vector but the form vector:
$$ \hat{\bm{f}}_M = M(\bm{\mathrm{f}}) $$
The motivator has the same architecture as the discriminator, and the same loss function. 
%
While the adversarial loss forces the encoder $E$ to produce a meaning vector $\bm{\mathrm{m}}$ with no information about the form $\bm{\mathrm{f}}$, the motivational loss encourages $E$ to encode this information in the form vector by minimizing $\mathcal{L}_M$:
$$ \mathcal{L}_\text{motiv}(\bm{\theta}_E) = \lambda_\text{motiv} \mathcal{L}_M $$

\subsection{Training procedure}
The overall training procedure follows the methods for training  GANs~\citep{goodfellow2014generative,arjovsky2017wasserstein} and consists of two stages: training the discriminator $D$ and the motivator $M$, and training the encoder $E$ and the generator $G$. 

In contrast to~\citet{arjovsky2017wasserstein}, we do not train the $D$ and $M$ more than the $E$ and the $G$. In our experiments we found that simple training in two stages is enough to achieve dissociation of the meaning and the form. Encoder and generator are trained with the following loss function that combines reconstruction loss with the losses from the discriminator and the motivator:
$$ \mathcal{L}_\text{total}(\theta_E,\theta_G) = \mathcal{L}_\text{rec} + \mathcal{L}_\text{adv} + \mathcal{L}_\text{motiv} $$

\section{Experimental setup}

\begin{figure*}[ht]
    \centering
  \begin{subfigure}[b]{0.45\textwidth}
        \centering
\includegraphics[width=1\textwidth]{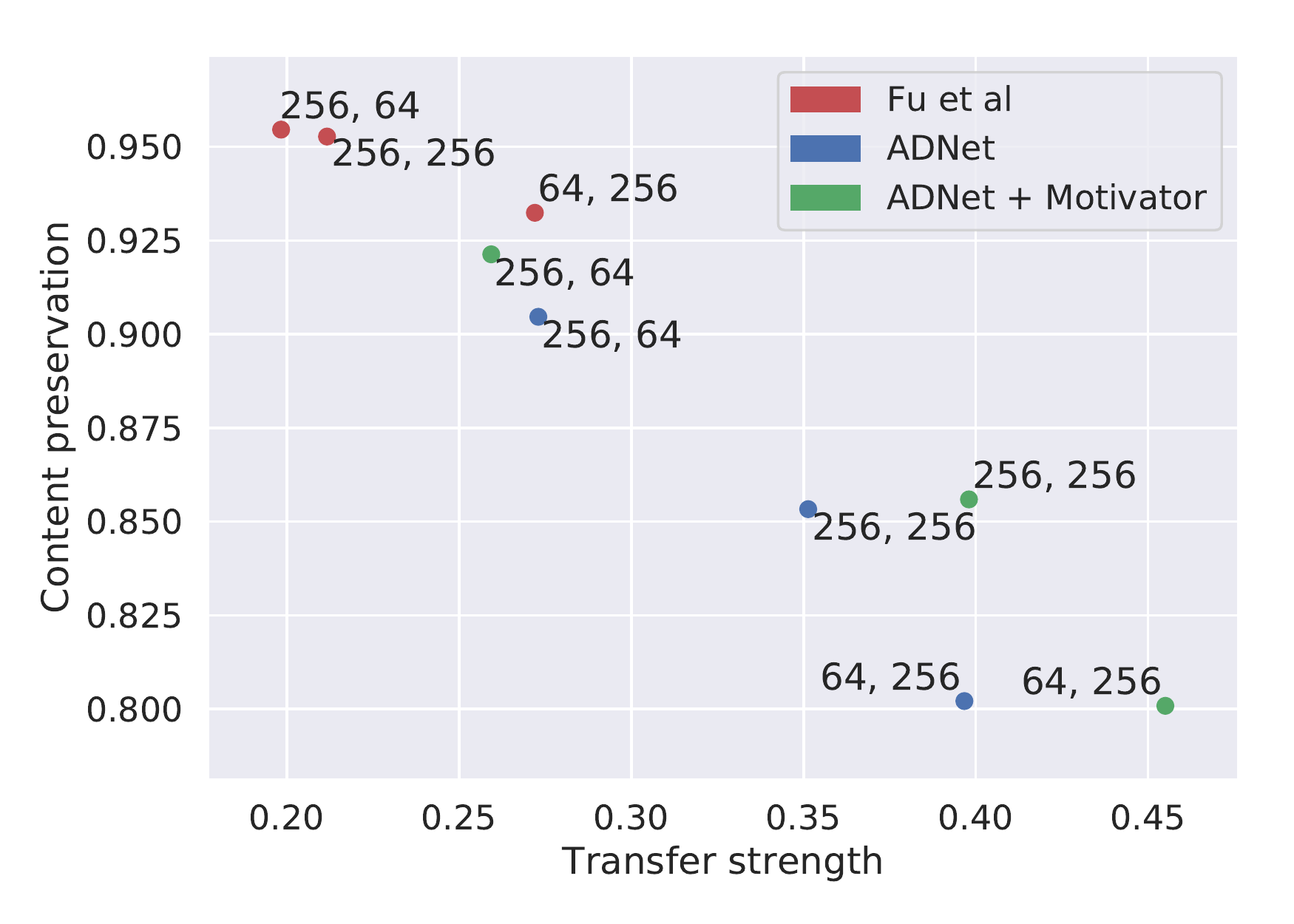}
    \caption{Shakespeare dataset}
    \label{fig:eval_shakespeare}  
    \end{subfigure}
  \begin{subfigure}[b]{0.45\textwidth}
    \centering
    \includegraphics[width=1\textwidth]{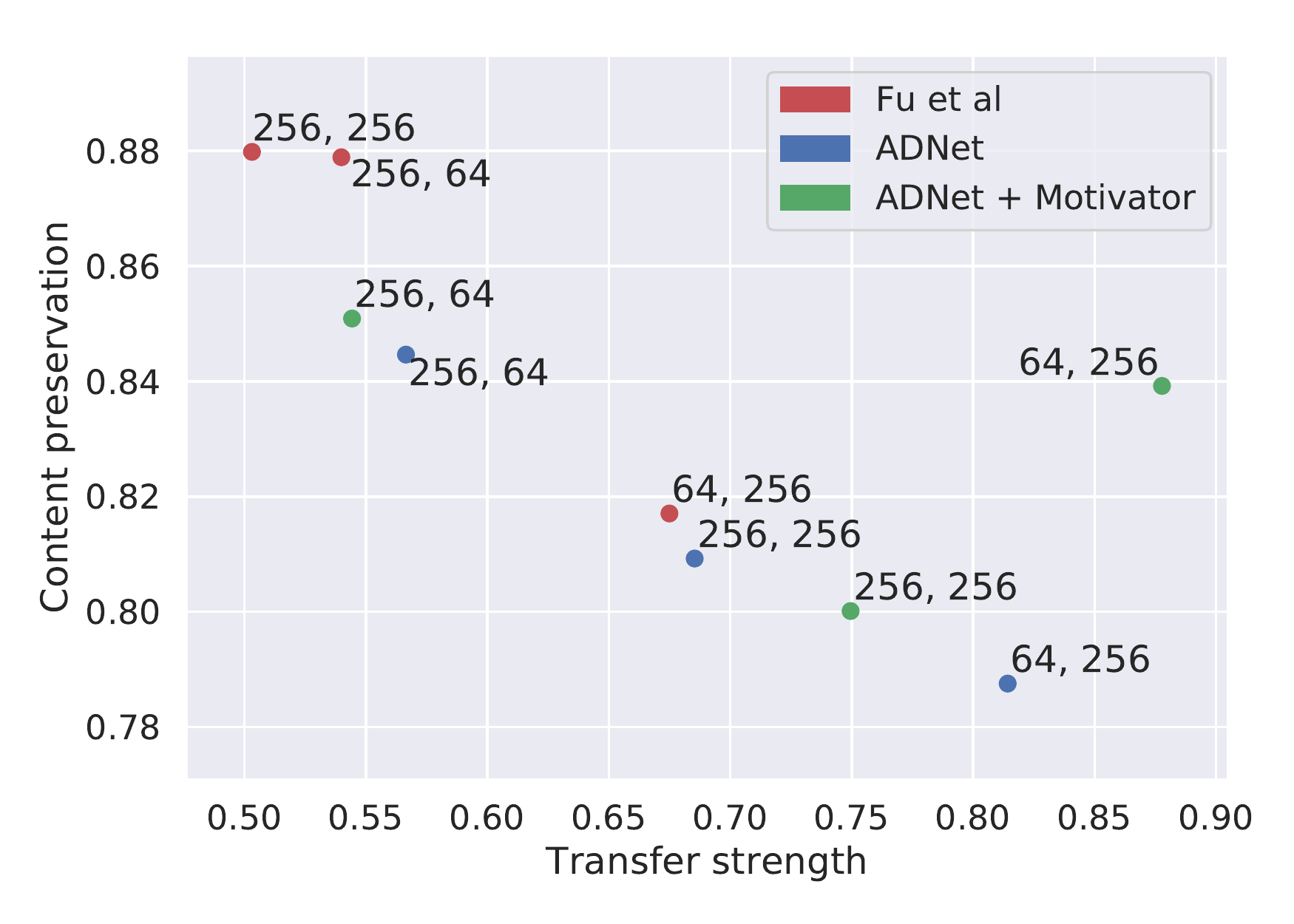}
    \caption{Headlines dataset}
    \label{fig:eval_headlines}  
    \end{subfigure}
  \caption{Transfer strength vs content preservation (see~\autoref{sec:evaluation}) for different sizes of the meaning and form vectors. Each point is labeled with ``\textlangle meaning vector size\textrangle , \textlangle form vector size\textrangle ''.}
    \label{fig:eval}  
\end{figure*}

\subsection{Evaluation}
\label{sec:evaluation}
Similarly to the evaluation of style transfer in CV~\citep{isola2017image}, evaluation of this task is difficult. We follow the approach of~\citet{isola2017image,shen2017style} and recently proposed by~\citet{fu2018style} methods of evaluation of ``transfer strength'' and ``content preservation''. The authors showed that the proposed automatic metrics correlate with human judgment to a large degree and can serve as a proxy. 
Below we give an overview of these metrics.

\paragraph{Transfer Strength.} The goal of this metric is to capture whether the form has been changed successfully. To do that, a classifier $C$ is trained on the two corpora, $\bm{X}^a$ and $\bm{X}^b$ to recognize the linguistic ``form'' typical of each of them. After that a sentence, for which the form/meaning has been changed, is passed to the classifier. The overall accuracy reflects the degree of success of changing the form/meaning. This approach is widely used in CV~\citep{isola2017image}, and was applied in NLP as well~\citep{shen2017style}.

In our experiments we used a GRU unit followed by four fully-connected layers with \verb|ELU| activation functions between them as the classifier.


\paragraph{Content preservation} Note that the transfer strength by itself does not capture the overall quality of a changed sentence. A extremely overfitted model that produces the most characteristic sentence of one corpus all the time 
would have a high score according to this metric. Thus, we need to measure how much of the meaning was preserved while changing the form. To do that, \citet{fu2018style} proposed to use a cosine similarity based metric using pretrained word embeddings. First, a sentence embedding is computed by concatenation of max, mean, and average pooling over the timesteps:
$$
\Resize{\linewidth}{
   \bm{v} = [\max(\bm{v}_1,\dotsc,\bm{v}_T);\min(\bm{v}_1,\dotsc,\bm{v}_T);\mean(\bm{v}_1,\dotsc,\bm{v}_T)] 
   }
$$

Next, the cosine similarity score $s_i$ between the embedding $\bm{v}^s_i$ of the original source sentence and the target sentence with the changed form $\bm{v}^t_i$ is computed, 
and the scores across the dataset are averaged to obtain the total score $s$.

\subsubsection{Continuous form}
\label{sec:continuous_form}

The metrics described above treat the form as a categorical (in most cases, even binary) variable. This was not a problem in previous work since the change of form could be done by simply inverting the form vector. Since we treat the form as a continuous variable, we cannot just use the proposed metrics directly. To enable a fair comparison, we propose the following procedure. 

For each sentence $s^a_s$ in the test set from the corpus $\bm{X}^a$ we sample $k = 10$ random sentences from the corpus $\bm{X}^b$ of the opposite form. After that, we encode them into the meaning $\bm{m}_i$ and form $\bm{f}_i$ vectors, and average the form vectors to obtain a single form vector 
$$\bm{f}_\text{avg} = \frac{1}{k}\sum_{i=1}^{k} \bm{f}_i$$
We then generate a new sentence with its original meaning  vector $\bm{m}_s$ and the resulting form vector $\bm{f}_\text{avg}$, and use it for evaluation. This process enables a fair comparison with the previous approaches that treat form as a binary variable.

\subsection{Datasets}

We evaluated the proposed method on several datasets that reflect different changes of meaning and form.

\paragraph{Changing form: register.}

This experiment is conducted with a dataset of titles of scientific papers and news articles published by~\citet{fu2018style}. This dataset (referred to as ``Headlines'') contains titles of scientific articles crawled from online digital libraries, such as ``ACM Digital Library'' and ``arXiv''. The titles of the news articles are taken from the ``News Aggregator Data Set'' from UCI Machine Learning Repository~\citep{Dua:2017} 

\paragraph{Changing form: language diachrony.}

Diachronic language change is explored with the dataset composed by~\citet{xu2012paraphrasing}. It includes the texts of 17 plays by William Shakespeare in the original Early Modern English, and their translations into contemporary English. We randomly permuted all sentences from all plays and sampled the training, validation, and test sets. Note that this dataset is much smaller than the Headlines dataset.







\section{Results and discussion}

\begin{figure*}[ht]
    \centering
  \begin{subfigure}[b]{0.48\textwidth}
        \centering
\includegraphics[width=1\textwidth]{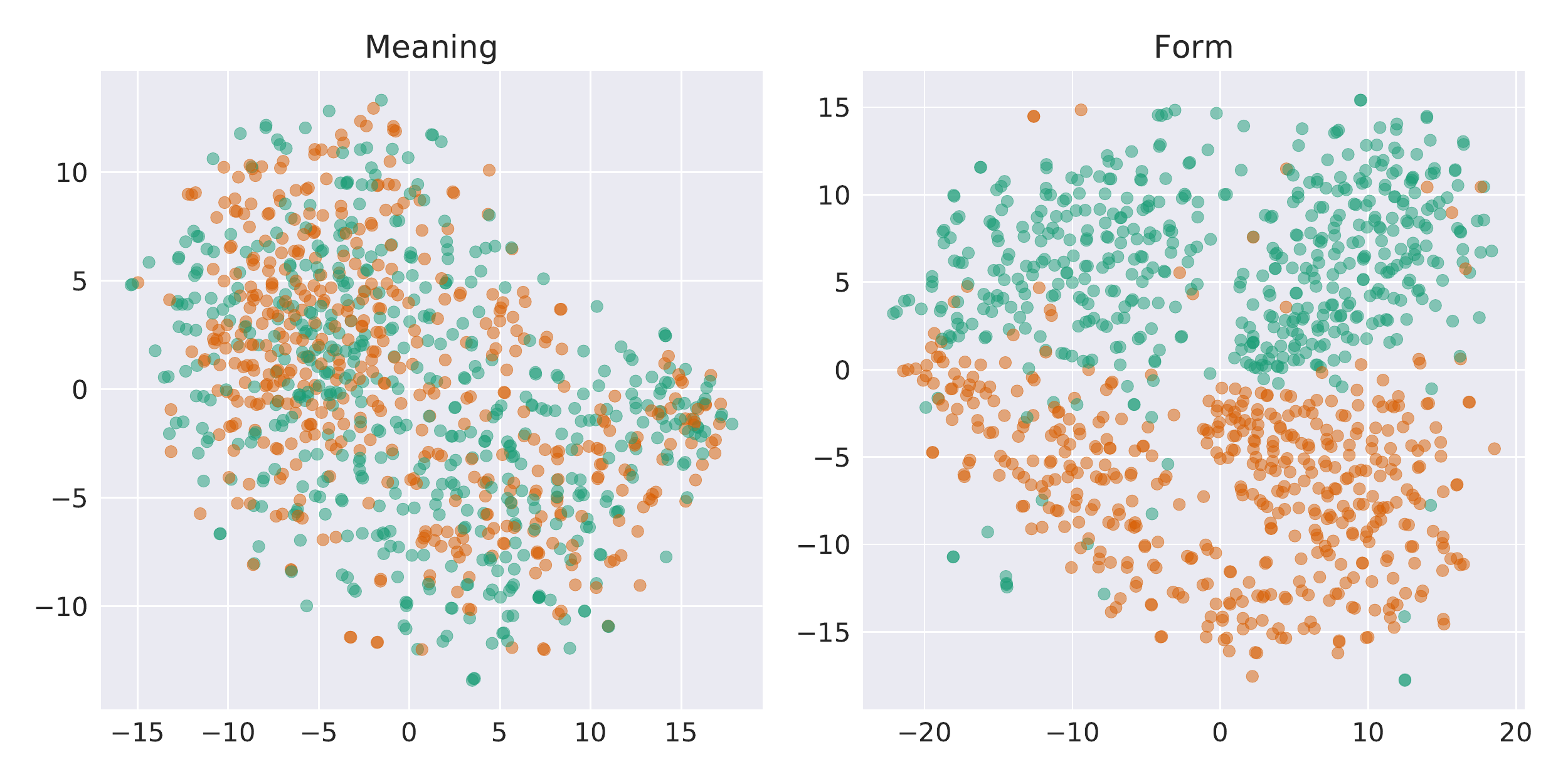}
    \caption{Without motivator}
    \label{fig:tsne_headlines_no_motivator}  
    \end{subfigure}
  \begin{subfigure}[b]{0.48\textwidth}
    \centering
    \includegraphics[width=1\textwidth]{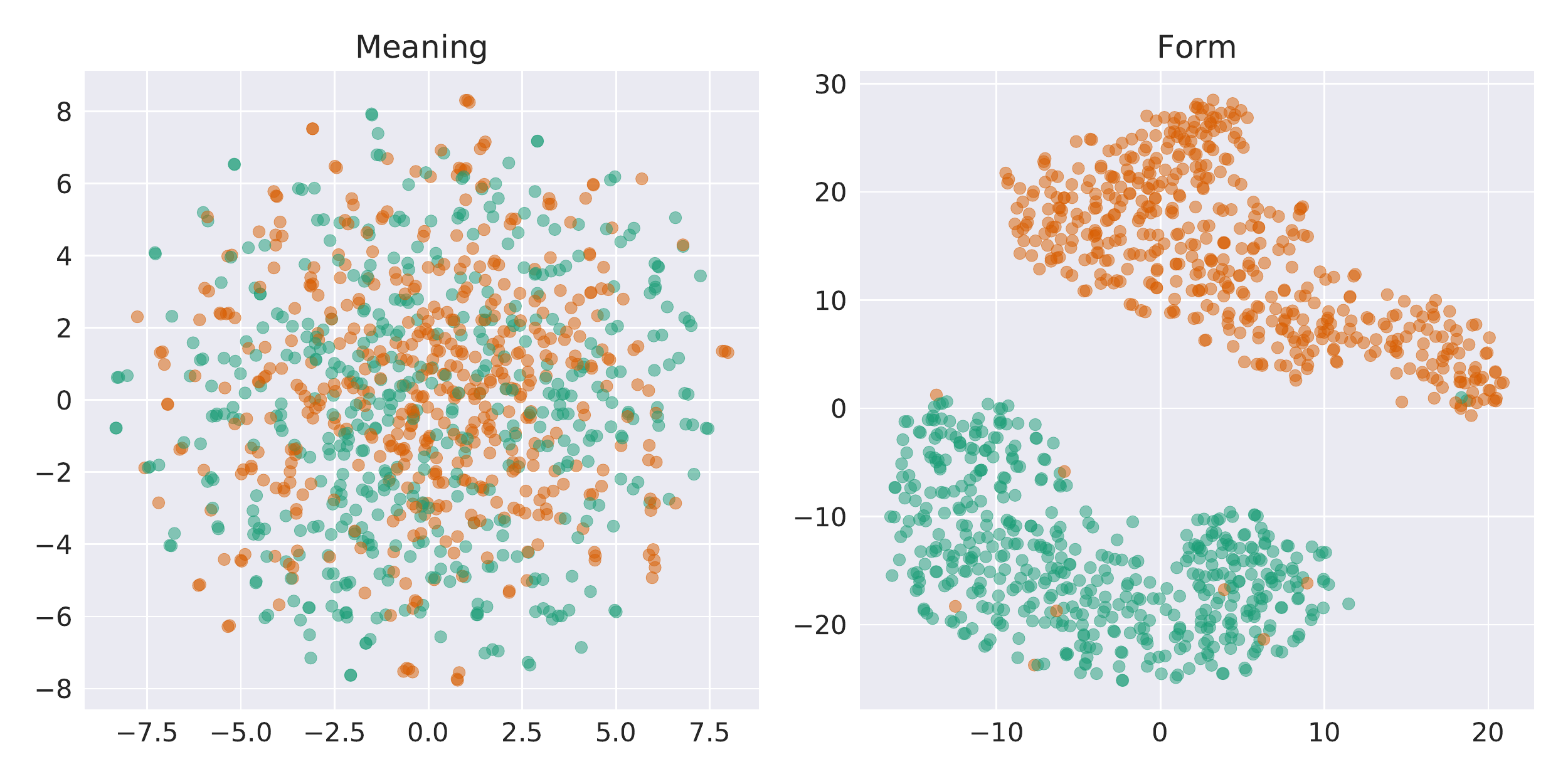}
    \caption{With motivator}
    \label{fig:tsne_headlines_motivator}  
    \end{subfigure}
  \caption{t-SNE visualization of the form and meaning embeddings of 1000 random sentences. Green point represent sentences form news headlines, and red points represent titles of scientific articles.}
    \label{fig:tsne_headlines}  
\end{figure*}

The most recent and similar to our work is the model proposed by~\citet{fu2018style}, in particular the  ``style-embedding'' model. We implemented this model to provide a baseline for comparison. 

The classifier used in the transfer strength metric achieves high accuracy (0.832 and 0.99 for the Shakespeare and Headlines datasets correspondingly). These results concur with the results of~\citet{shen2017style} and~\citet{fu2018style}, and show that the two corpora are significantly different. 


Following~\citet{fu2018style}, we show the result of different configuration of the size of the form and meaning vectors on~\autoref{fig:eval}. Namely, we report combinations of 64 and 256-dimensional vectors. Note that the sizes of the form vector are important. If the form vector is larger, the transfer strength is gre,ta erbut the content preservation is lessened. This is consistent with~\citet{fu2018style}, where they observed a similar behaviour. 

It is clear that the proposed method achieves significantly better transfer strength than the previously proposed model. It also has a lower content preservation score, which means that it repeats fewer exact words from the source sentence. Note that a low transfer strength and very high (\texttildelow 0.9) content preservation score means that the model was not able to successfully learn to transfer the form and the target sentence is almost identical to the source sentence. The Shakespeare dataset is the hardest for the model in terms of transfer strength, probably because it is the smallest dataset, but the proposed method performs consistently well in transfer of both form and meaning and, in contrast to the baseline.

\paragraph{Storing meaning in the form vector}
Note that, theoretically, nothing is stopping the model from storing the meaning in the form vector, except from the size limitations, which would ensure that storing non-form-related information elsewhere would improve model performance. \autoref{fig:eval} shows that as the meaning vectors get smaller, and the form vectors larger, the higher is transfer strength and the lower is content preservation. If the model would store meaning in the form vector, then the reduction in size of the meaning vector would not have negative impact on content preservation. This shows that the model tends to not store the meaning in the form vector.

Nevertheless, to force this behaviour we experimented with adding one more discriminator $D_f$. This discriminator works on the form vector $\bm{\mathrm{f}}$ in the same manner as the discriminator $D$ works on the meaning vector $\bm{\mathrm{m}}$. Namely, during the training it tries to predict the meaning of a sentence from its form vector: ${\bm{u}} = D_f(\bm{\mathrm{f}})$. Note that the vectors $\bm{u}$ and $\bm{\mathrm{m}}$ are completely different. $\bm{\mathrm{m}}$ is the meaning of a sentence for the  purpose of the model, whereas $\bm{u}$ are pre-defined meaning of a sentence for training of the discriminator. In the simplest case, $\bm{u}$ can be a multi-hot representation of the input sentence, with the exception of pre-defined ``style'' words, which would always have $0$ in the corresponding dimension, as it is done by~\newcite{john2018disentangled}. 

We, however, take a different approach. First, we find the ``form'' dimensions in the used word embeddings by taking the argmax of the difference between averaged word embeddings of the sentences from two forms (i.e. Early Modern English and contemporary English). Next, for a given sentence we discard the top-$k$ tokens with the maximum and minimum values in those dimensions. Finally, we average word embeddings of the remaining tokens in the sentence to get the vector $\bm{u}$. Such incorporation of the discriminator $D_f$ helped to mitigate this issue.

\paragraph{Fluency of generated sentences} Note that there is no guarantee that the generated sentences would be coherent after switching the form vector. In order to estimate how this switch affects the fluency of generated sentences, we trained a language model on the Shakespeare dataset and calculated the perplexity of the generated sentences using the original form vector and the average of form vectors of $k$ random sentences from the opposite form (see~\autoref{sec:continuous_form}). While the perplexity of such sentences does go up, this change is not big (6.89 vs 9.74).

\subsection{Impact of the motivational training}

\begin{table*}
    \centering
\resizebox{0.98\linewidth}{!}{
\begin{tabular}[t]{@{}lll@{}}
\toprule
Aye, sir. (EME)  & \textrightarrow & Yes, sir. (CE) \\
Fare thee well, my lord (EME) & \textrightarrow & Fare you well, my lord (CE) \\
This guy will tell us everything. (CE) & \textrightarrow & This man will tell us everything. (EME) \\
I've done no more to caesar than you will do to me. (CE) & \textrightarrow & I have done no more to caesar than, you shall do to me. (EME)  \\ \bottomrule
\end{tabular}
}
\caption{Decoding of the source sentence from Early Modern English (EME) into contemporary English (CE), and vice versa.}
\label{tbl:shakespeare_sentences}

\end{table*}

\begin{table*}[!ht]
\centering
\resizebox{0.98\linewidth}{!}{
\begin{tabular}{@{}lp{1cm}l@{}}
\toprule
A review: detection techniques for LTE system & & Crisis management: media practices in telecommunication management \\
Situation management knowledge from social media & & A review study against intelligence internet \\ \midrule
Security flaw could not affect digital devices, experts say & & Semantic approach approach: current multimedia networks as modeling processes \\
Semantic approach to event processing & & Security flaw to verify leaks \\
\bottomrule
\end{tabular}
} 

\begin{tikzpicture}[overlay]
\tikzmath{
    \xbase = -1.7; \ybase = 2.05; \xoffset = 0.8; \yoffset = -0.26; \yspace = -0.22;
    \x1 = \xbase; 
    \x2 = \xbase + 1 * \xoffset;
    \y1 = \ybase; 
    \y2 = \ybase + 1 * \yoffset; 
    \y3 = \ybase + 2 * \yoffset + 1 * \yspace;
    \y4 = \ybase + 3 * \yoffset + 1 * \yspace;
    \y5 = \ybase + 4 * \yoffset + 2 * \yspace;
    \y6 = \ybase + 5 * \yoffset + 2 * \yspace;
} 
\draw [->,out=0,in=180,line width=0.7] (\x1, \y3) to (\x2, \y4);
\draw [->,out=0,in=180,line width=0.7] (\x1, \y4) to (\x2, \y3);
\draw [->,out=0,in=180,line width=0.7] (\x1, \y5) to (\x2, \y6);
\draw [->,out=0,in=180,line width=0.7] (\x1, \y6) to (\x2, \y5);
\end{tikzpicture}

\caption{Flipping the meaning and the form embeddings of two sentence from different registers. Note the use of colon in the first example, and the use of the ``to''-constructions in the second example, consistent with the form of the source sentences.}
\label{tbl:headlines_sentence}
\end{table*}

To investigate the impact of the motivator, we visualized form and meaning embeddings of 1000 random samples from the Headlines dataset using t-SNE algorithm~\citep{van2014accelerating} with the Multicore-TSNE library~\citep{Ulyanov2016}. The result is presented in~\autoref{fig:tsne_headlines}. 

There are three important observations. 
First, there is no clear separation in the meaning embeddings, which means that any accurate form transfer is due to the form embeddings, and the dissociation of form and meaning was successful.

Second, even without the motivator the model is able to produce the form embeddings that are clustered into two groups. Recall from~\autoref{sec:method_description} that without the motivational loss there are no forces that influence the form embeddings, but nevertheless the model learns to separate them. 

However, the separation effect is much more pronounced in the presence of motivator. This explains why the motivator consistently improved transfer strength of ADNet, as shown in \autoref{fig:eval}.

\subsection{Qualitative evaluation}
\label{sec:qualitative_eval}

\autoref{tbl:shakespeare_sentences} and~\autoref{tbl:headlines_sentence} show several examples of successful form/meaning transfer achieved by ADNet. \autoref{tbl:shakespeare_sentences} presents the results of an experiment that to some extent replicates the approach taken by the authors who treat linguistic form  as a binary variable~\citep{shen2017style,fu2018style}. The sentences the original Shakespeare plays were averaged to get the ``typical'' Early Modern English form vector. This averaged vector was used to decode a sentence from the modern English translation back into the original. The same was done in the opposite direction. 

\autoref{tbl:headlines_sentence} illustrates the possibilities of ADNet on fine-grained transfer applied to the change of register. We encoded two sentences in different registers from the Headlines dataset to produce form and meaning embeddings, and then decoded the first sentence with the meaning embedding of the second, and vice versa. \autoref{tbl:headlines_sentence} shows that the model correctly captures the meaning of sentences and decodes them using the form of the source sentences, preserving specific words and the structure of the source sentence. Note that in the first example, the model decided to put the colon after the ``crisis management'', as the source form sentence has this syntactic structure (``A review:'').  This is not possible in the previously proposed models, as they treat form as just a binary variable.


\subsection{Performance of meaning embeddings on downstream tasks}

\begin{table}[ht]
\centering
\resizebox{0.98\linewidth}{!}{
\begin{tabular}{@{}lllll@{}}
\toprule
BoW   & Seq2Seq & InferSent & \citet{fu2018style} & ADNet \\ \midrule
80.82 & 74.68   & \textbf{83.17}     & 78.88    & 81.38           \\ \bottomrule
\end{tabular}
}
\caption{F1 scores on the task of paraphrase detection using the SentEval toolkit~\citep{conneau2017supervised}}
\label{tbl:paraphrase_results}
\end{table}

We conducted some experiments to test the assumption that the derived meaning embeddings should improve performance on downstream tasks that require understanding of the meaning of the sentences regardless of their form. 
We evaluated embeddings produced by the ADNet, trained in the Headlines dataset, on the paraphrase detection task.
We used the SentEval toolkit~\citep{conneau2017supervised} and the Microsoft Research Paraphrase Corpus~\citep{dolan2004unsupervised}. 
%
The F1 scores on this task for different models are presented in~\autoref{tbl:paraphrase_results}. Note that all models, except InferSent, are unsupervised. The InferSent model was trained on a big SNLI dataset, consisting of more than 500,000 manually annotated pairs.
ADNet achieves the the highest score among the unsupervised systems and far outperforms the regular sequence-to-sequence autoencoder.

\begin{figure*}[ht]
    \centering
    \begin{subfigure}[b]{0.45\textwidth}
    \centering
    \includegraphics[width=0.5\textwidth]{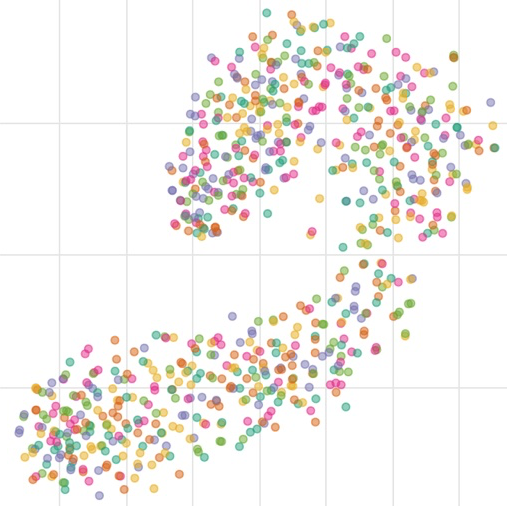}
    \caption{Meaning embeddings}
    \label{fig:tsne_authorship_meaning}  
    \end{subfigure}
  \begin{subfigure}[b]{0.45\textwidth}
    \centering
    \includegraphics[width=0.5\textwidth]{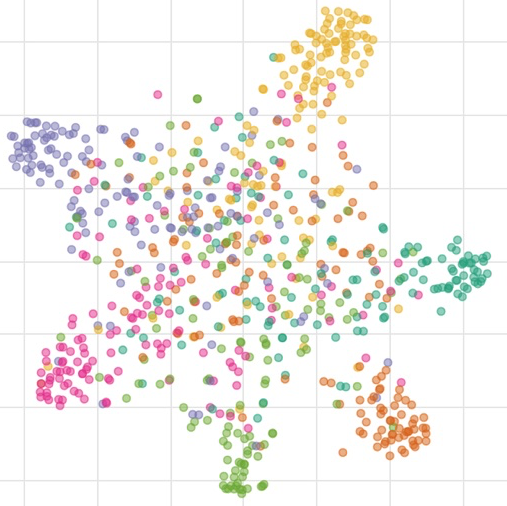}
    \caption{Form embeddings}
    \label{fig:tsne_authorship_form}  
    \end{subfigure}
  \caption{t-SNE visualization of the form and meaning embeddings. Each color corresponds to a different author.}
    \label{fig:tsne_authorship}  
\end{figure*}

\subsection{Multiple forms and stylistic similarities}
\label{sec:multiple_forms}

In order to go beyond just two different forms, we experimented with training the model on a set of literature novels from six different authors from Project Gutenberg\footnote{\url{http://www.gutenberg.org/}} written in two different time periods. 
A t-SNE visualization of the resulting meaning and form embeddings is presented in~\autoref{fig:tsne_authorship}. Note how form embeddings create a six-pointed star. After further examination, we observed that common phrases (for example, ``Good morning'' or ``Hello!'') were embedded into the center of the star, whereas the most specific sentences from a given author were placed into the rays of the star. In particular, some sentences included character names, thus further research is required to mitigate this problem. \citet{stamatatos2017authorship} provides a promising direction for solving this.

\section{Conclusion}
We presented ADNet, a new model that performs adversarial decomposition of text representation. In contrast to previous work, it does not require a parallel training corpus and works directly on hidden representations of sentences. Most importantly, it does not treat the form as a binary variable (as done in most previously proposed models), enabling a fine-grained change of the form of sentences or specific aspects of meaning. We evaluate ADNet on two tasks: the shift of language register and diachronic language change. Our solution achieves superior results, and t-SNE visualizations of the learned meaning and form embeddings illustrate that the proposed motivational loss leads to significantly better separation of the form embeddings.

\bibliographystyle{acl_natbib}
\bibliography{main}

\end{document}